\documentclass[letterpaper]{article} 

\usepackage{pdfpages} 

\usepackage{aaai2026}  
\usepackage{times}  
\usepackage{helvet}  
\usepackage{courier}  
\usepackage[hyphens]{url}  
\usepackage{graphicx} 
\usepackage{natbib}  
\usepackage{caption} 
\usepackage{amsmath}
\usepackage{amsfonts}
\usepackage{booktabs}
\usepackage{multirow}
\usepackage{array} 
\usepackage{algorithm}
\usepackage{algorithmic}

\usepackage{newfloat}
\usepackage{listings}
\DeclareCaptionStyle{ruled}{labelfont=normalfont,labelsep=colon,strut=off} 
\floatstyle{ruled}
\newfloat{listing}{tb}{lst}{}
\floatname{listing}{Listing}
\title{Reinforced Language Models for Sequential Decision Making}
\author{
    Jim Dilkes, Vahid Yazdanpanah, Sebastian Stein 
}
\affiliations{
    University of Southampton \\
    {j.dilkes@soton.ac.uk}
    
}

\begin{document}

\nocopyright
\maketitle

\begin{abstract}

Large Language Models (LLMs) show potential as sequential decision-making agents, but their application is often limited due to a reliance on large, computationally expensive models. This creates a need to improve smaller models, yet existing post-training methods are designed for single-turn interactions and cannot handle credit assignment in multi-step agentic tasks. To address this, we introduce Multi-Step Group-Relative Policy Optimization (MS-GRPO), a new algorithm for post-training LLM agents, grounded in formal Text-Mediated Stochastic Game (TSMG) and Language-Agent Policy (LAP) frameworks. For credit assignment, MS-GRPO attributes the entire cumulative episode reward to each individual episode step. We supplement this algorithm with a novel absolute-advantage-weighted episode sampling strategy that we show improves training performance. We evaluate our approach by post-training a 3-billion parameter model on Snake and Frozen Lake. Our experiments demonstrate that the method is effective in improving decision-making performance: our post-trained 3B parameter model outperforms a 72B parameter baseline by $50\%$ on the Frozen Lake task.
This work demonstrates that targeted post-training is a practical and efficient alternative to relying on model scale for creating sequential decision-making agents using LLMs.

\end{abstract}

%

\section{Introduction}

Sequential decision making, the problem of an agent selecting successive actions to maximize a long-term objective, represents a fundamental and pervasive challenge in artificial intelligence. Computational approaches to this problem have driven significant achievements in applications as diverse as spacecraft control \citep{bernard_design_1998}, medical treatment \citep{murphy_optimal_2003,bani-harouni_language_2025}, robotic manipulation \citep{levine_end--end_2016}, data center cooling efficiency \citep{evans_deepmind_2016}, and vehicle routing \citep{kool_attention_2019}. Recently, the powerful reasoning and natural language understanding capabilities of Large Language Models (LLMs) have enabled a new paradigm: agents that can follow human instruction to operate in dynamic environments conveyed through text, whether digital \citep{zheng_deepresearcher_2025} or physical \citep{mower_ros-llm_2024, li_hmcf_2025}. These hold great potential in leveraging the extensive world knowledge and reasoning abilities inherent in LLMs to flexibly tackle sequential decision-making problems.

Despite this promise, effectively utilizing LLMs for sequential decision making remains an open challenge. 
Specifically, evidence suggests that LLMs struggle with low-level action selection \citep{zhang_building_2024}, are not inherently good planners \citep{kambhampati_position_2024}, and effective decision making typically requires large models using computationally expensive reasoning chains \citep{tanahashi_evaluation_2023, yao_react_2023-1, shinn_reflexion_2023, zhou_reflect-rl_2024}. For example,  \citet{trivedi_appworld_2024} find that their most capable agent, using GPT-4o \citep{openai_gpt-4o_2024} on realistic digital tasks,
costs $\$0.70$ per task while achieving less that $50\%$ success rate.  These shortcomings limit the practicality and scalability of LLMs, highlighting the need for new training methods to enhance the capabilities of more efficient models.

However, existing LLM post-training methods, those that refine and adapt a pre-trained model to meet application-specific requirements, are unsuitable for this domain. These approaches, often based on Reinforcement Learning (RL) \citep{sutton_reinforcement_2018}, are designed to optimize models on single-turn tasks with immediate feedback from a verifier, as in Reinforcement Learning with Verifiable Rewards (RLVR) \citep{deepseek-ai_deepseek-r1_2025, zheng_group_2025, yu_dapo_2025,wang_reinforcement_2025, hou_t1_2025, park_maporl_2025}, or from human preference models, as in Reinforcement Learning from Human Feedback (RLHF) \citep{ziegler_fine-tuning_2020, ouyang_training_2022, rafailov_direct_2023, zhong_dpo_2025}. Such methods, however, are incompatible with sequential decision-making tasks where credit assignment of outcomes to actions is necessary. Addressing this is an emerging research area. For example, the RAGEN system \citep{wang_ragen_2025} conditions the agent's language generation on full environment episodes, assigning credit for the entire episode to the agent's complete sequence of actions.


Furthermore, a conceptual limitation arises when using LLMs as decision-making agents: the optimization occurs over sequences of \textit{tokens}, which are communicative units rooted in natural language, whereas effective planning requires the selection of \textit{actions} grounded in the problem domain (e.g. navigation moves in a spatial environment). This discrepancy mirrors the distinction between communicative acts, such as speech acts in dialogue systems \cite{traum_speech_1999}, and operational actions needed for sequential decision-making \cite{georgeff_theory_1988}. Bridging this gap calls for new methods that formally align the language-centric outputs of LLMs with the structured, domain-specific actions required for agent planning and control.

Against this background, for the first time, we:
\begin{enumerate}
    \item Define a formal framework connecting language-based agents and sequential decision-making environments, comprising the Text-Mediated Stochastic Game (TMSG), which models the environment with an explicit text interface, and Language Agent Policy (LAP), which defines the agent's LLM-based policy.
    \item Introduce Multi-Step Group-Relative Policy Optimization (MS-GRPO), an algorithm adapting the GRPO method for sequential decision-making tasks by assigning the entire cumulative episode reward to each individual step. To improve efficiency, the optimization for each step uses only the current state as context.
    \item Propose a novel absolute-advantage-weighted (AAW) episode sampling strategy which we demonstrate improves training performance.
    \item Demonstrate that our post-trained 3B parameter model outperforms a much larger 72B parameter baseline LLM on the Frozen Lake task by $50\%$, showing the value of domain-specific training over model scale.
    \item Provide a critical analysis of the MS-GRPO algorithm's capabilities, highlighting its high training variance and mixed results in eliciting generalization in LLM-based agents. 
\end{enumerate}

\section{Framework}
\label{sec:framework}
This section defines our framework for language model-based agents in sequential decision-making environments. The framework consists of two core contributions: 
(1) a Text-Mediated Stochastic Game (TMSG), that formalizes an environment where all interactions are mediated exclusively through text; and
(2) a Language Agent Policy (LAP) that parameterizes an agent's behavior in terms of a language model and interfacing components.
Our framework's explicit separation of agent and environment is important for two reasons. First, it allows us to model the TMSG using the formalisms of Partially-Observable Stochastic Games (POSG) \citep{hansen_dynamic_2004}. 
Second, the LAP formalism makes explicit which components we can control to affect the agent's decision making. 

\subsection{Text-Mediated Stochastic Game}

To formally model sequential decision-making environments with text-based interfaces, we define a Text-Mediated Stochastic Game (TMSG) as a tuple $G = \left(\mathcal{P}, \mathcal{S}, \mathcal{A}, \Omega, O, P, R\right)$. This formalism builds on Stochastic Games \citep{shapley_stochastic_1953} and has similarities to Partially-Observable Stochastic Games, with the key constraint that observation space for each agent is the set of all text strings. While POSGs provide a familiar structure, the TMSG makes the text interface between an LLM-based agent and the environment explicit. The components of $G$ are defined as follows:

\begin{itemize}
    \item $\mathcal{P} = \{1, \ldots, p\}$ is a finite set of $p$ players.\footnote{While this work focuses on single-agent post-training, we adopt the more general multi-agent case to provide a robust framework. Non-learning agents, such as the opponent in the Snake environment, are therefore treated as a component of the environment and are not the subject of this work.}
    
    \item $\mathcal{S}$ is the finite set of game states. The set of terminal states is denoted $\mathcal{S}_f \subset \mathcal{S}$.
    
    \item $\mathcal{A} = \mathcal{A}_1 \times \dots\times\mathcal{A}_p$ is the joint action space where $\mathcal{A}_i$ is the set of all possible actions for agent $i$. The function $\mathcal{A}(s)\subseteq\mathcal{A}$ returns the set of legal joint actions in state $s$.

    \item $\Omega = \Omega_1\times\dots\times\Omega_p$ is the joint observation space, where each agent-specific observation space $\Omega_i=\Sigma^*$ is the set of all text strings $\Sigma^*$, with $\Sigma$ the vocabulary of tokens.

    \item An observation function $O:\mathcal{S}\times \mathcal{A} \rightarrow \Delta(\Omega)$, where $O(o|s',a)$ is the probability of the joint observation $o$ after taking joint action $a$ and transitioning to state $s'$.
    
    \item $P: \mathcal{S} \times \mathcal{A} \rightarrow \Delta (\mathcal{S})$ is a state transition function where $P(s'|s,a)$ is the probability of transitioning to state $s'$ when the joint action $a$ is taken in state $s$.
    
    \item $R: \mathcal{S} \times \mathcal{A} \rightarrow \mathbb{R}^p$ is the reward function, where $R(s,a)$ returns a list of $p$ numerical values indicating the reward for each player after taking the joint action $a$ in state $s$.
\end{itemize}

\subsection{Language Agent Policy}

In the settings we consider, a player $i\in P$ is controlled by a computational agent whose goal is to maximize its expected cumulative reward. To achieve this, the agent learns a policy, which specifies a strategy for choosing actions based on its observations. Here we outline our proposed architecture for language-model-based agents, which we call a Language Agent Policy (LAP).

Previous research in RLVR considers the policy to be the probability distribution over tokens described by an LLM \citep{zheng_group_2025}. While this framing is useful, our focus on sequential tasks requires optimizing for environment actions, not only the tokens that produce them.

An agent's policy specifies the probability of it taking action $a$ given observation $o$. In our LLM-based framework the policy is parameterized by a set of components $\Pi_i=(\mathcal{L}_{\theta_i}, \mathcal{G}_i, \mathcal{T}_i, \psi_i)$ comprising:
\begin{itemize}
    \item $\mathcal{L}_{\theta_i}$: a generative language model with parameters $\theta_i$.
    \item $\mathcal{G}_i$: the generation configuration (e.g., temperature, top-k) controlling the token sampling behavior of $\mathcal{L}_{\theta_i}$. 
    \item $\mathcal{T}_i$: a prompt template that is a text string with a placeholder to be filled with an observation string to create a full input prompt.
    \item $\psi_i:  \Sigma^* \rightarrow \mathcal{A}_i \cup \{\perp\}$: an action extraction function that parses the text output of $\mathcal{L}_{\theta_i}$ and maps it to a valid game action, or to $\perp$ if the output cannot be interpreted as a valid action.
\end{itemize}
The parameterized policy can be written as $\pi_{\Pi_i}(a|o)\in[0,1]$. We denote the language model $\mathcal{L}_{\theta_i}$ operating under a specific generation configuration $\mathcal{G}_i$ as $\mathcal{L}_{\theta_i|\mathcal{G}_i}$. This symbol represents the resulting stochastic text generation function. 

An action $a_t^i$ is sampled from the language agent policy at time $t$ through the following process:
\begin{enumerate}
    \item \textbf{Prompt Construction:} A prompt $q_t^i \in \Sigma^*$ is constructed from the observation $o_t^i$ using the agent's template 
    $q_t^i=\mathcal{T}_i(o_t^i)$. 
    \item \textbf{Stochastic Text Generation:} A text completion, $c_t^i \in \Sigma^*$, is sampled from the language model given the input prompt $q_t^i$:
        $c_t^i \sim \mathcal{L}_{\theta_i|\mathcal{G}_i}\left(\cdot | q_t^i\right)$.
    \item \textbf{Action Parsing:} The action taken by the agent $a_t^i$ is extracted from the completion string by the parsing function
        $a_t^i=\psi_i(c_t^i)$.
\end{enumerate}

\subsection{Agent-Environment Interaction}
The interaction between LAP agents and the TMSG environment proceeds in discrete time steps. This agent-environment loop applies to any set of agents, each implementing a policy $\pi_i$. In this work, we are primarily interested in the case where the policy is a LAP, $\pi_{\Pi_i}$. 

The sequence of events at each time step $t$, starting from an initial state $s_0$, is as follows:

\begin{enumerate}
    \item Each player $i \in P$ simultaneously selects an action: 
    $a_t^i \sim \pi_i(\cdot | o_t^i)$. If $a_t^i = \perp$, apply a predefined recovery strategy (e.g.,  no action or a random action). The collection of actions from all agents forms the joint action $a_t = (a_t^1, \ldots, a_t^p)$.
    \item The environment receives joint action $a_t$ and transitions from state $s_t$ to state $s_{t+1}$ by sampling from the state transition function: 
    $s_{t+1}\sim P(\cdot| s_t,a_t)$. 
    \item The environment generates a vector of rewards $r_{t+1}=\left(r_{t+1}^1,\dots,r_{t+1}^p\right)$, calculated by the reward function $R(s_t,a_t)$.
    \item The environment samples a joint observation $o_{t+1}=(o_{t+1}^1,\dots,o_{t+1}^p)$ from the observation function: 
    $o_{t+1}\sim O(\cdot|s_{t+1},a_t)$. 
    \item If $s_{t+1} \in \mathcal{S}_f$, the episode terminates. Otherwise, increment $T$ and repeat.
\end{enumerate}

Together, the TMSG and LAP formalisms provide a complete framework for analyzing LLM agents in sequential decision-making environments.

\section{Methodology}
\label{sec:methodology}
Creating effective LLM-based agents for sequential decision making requires new methods that overcome the limitations of single-turn optimization algorithms, particularly the problem of credit assignment from sparse, delayed rewards. This section presents our methodology for solving this problem by training a goal-seeking LAP within the TMSG framework, which consists of two technical contributions:
\begin{enumerate}
    \item Multi-Step Group-Relative Policy Optimization (MS-GRPO): a new algorithm that specifies reward assignment from environment steps to tokens in order to handle multi-step trajectories. 
    \item Absolute-Advantage-Weighted Episode Sampling: an episode sampling strategy that prioritizes episodes with more extreme outcomes. 
\end{enumerate}

\subsection{GRPO Modification for Multi-Step Environments}
To optimize the behavior of LAP agents to maximize expected cumulative reward in TMSG environments we propose Multi-Step Group-Relative Policy Optimization (MS-GRPO). Our algorithm is a variation of GRPO \citep{shao_deepseekmath_2024}, a type of policy gradient method \citep{williams_simple_1992}.

Whereas GRPO compares rewards from single-step responses to an identical prompt, MS-GRPO adapts this approach for multi-step tasks. It calculates an advantage value from the total cumulative reward and assigns this value to every generated token in that episode. This technique of attributing the full episodic reward to each action is a form of Monte Carlo credit assignment \citep{sutton_reinforcement_2018}. We use GRPO instead of an actor-critic method such as PPO \citep{schulman_proximal_2017} due to its reduced memory footprint which allows training of larger models or the use of longer contexts.

While the LAP agent's behavior is defined by the complete policy $\pi_{\Pi}$, we optimize only the parameters $\theta$ that determine its LLM's distribution over tokens, $\mathcal{L}_{\theta|\mathcal{G}}\left(\cdot | q_t\right)$. Although the TMSG framework supports multiple players, this work focuses on optimizing a single agent, so we omit the player index $i$ in the following definition.

The MS-GRPO objective function is defined as:
\begin{multline}
\mathcal{J}_{\text{MS-GRPO}}(\theta) = 
\mathbb{E}_{o\sim\mathcal{D},\{y_{j,t}\}_{j=1}^G\sim p_{\theta_\text{old}}(\cdot|o_t)} \\
\frac{1}{G} \sum_{j=1}^G \frac{1}{|y_{j}|} \sum_{t=0}^{T_j-1}
\left[ \mathcal{L}_\text{CLIP}(\theta,j,t) \right] 
 - \beta \mathbb{D}_{\text{KL}}\left(p_\theta \| p_{\text{ref}}\right)
\end{multline}
where $\mathcal{L}_\text{CLIP}$ is the token-level objective for timestep $t$ in episode $j$:
\begin{multline}
    \mathcal{L}_\text{CLIP}(\theta,j,t) = \\
    \sum_{k=1}^{|y_{j,t}|} \min \left( w_{j,t,k}A_j, \right.
\left.\left.\left. \text{clip} \left(w_{j,t,k}, 1-\epsilon_\text{low}, 1+\epsilon_\text{up} \right)A_j \right) \right.\right.
\end{multline}
and the importance ratio $w_{j,t,k}$ is:
\begin{equation}
    w_{j,t,k} = 
    \frac{p_\theta(y_{j,t,k} \mid o_{j,t}, y_{j,t,<k})}{p_{\theta_\text{old}}(y_{j,t,k} \mid o_{j,t}, y_{j,t,<k})}
\end{equation}

Here, $G$ is the group size, $T_j$ is the number of timesteps in episode $j$, $|y_j|$ is the total number of generated tokens, and $y_{i,t,k}$ is $k$-th token in the completion at timestep $t$. $\mathcal{D}$ denotes the distribution over observations determined by the TMSG dynamics. The episode advantage, $A_j$, is calculated by normalizing a composite reward $C_j=\mathcal\sum_{T_j}(r_{j,t} +\Phi_{j,t})$, which combines the cumulative environment reward
with a task-specific shaping reward ($\Phi_j$):
\begin{equation}
\label{eqn:advantage}
A_j = \frac{C_j - \text{mean}(\{C_1, \dots, C_G\})}{\text{std}(\{C_1, \dots, C_G\})}
\end{equation}
Finally, $\mathbb{D}_{\text{KL}}\left(p_\theta \| p_{\text{ref}}\right)$ is the KL penalty against a reference model (the original LLM before post-training), and $\epsilon_\text{low}$, $\epsilon_\text{up}$ and $\beta$ are hyperparameters. 
The MS-GRPO algorithm is detailed in Algorithm \ref{alg:msgrpo}.

\begin{algorithm}[!htb]
\caption{Multi-Step Group Relative Policy Optimization for Language Agent Policies}
\begin{algorithmic}[1]
\label{alg:msgrpo}
\REQUIRE Initial model parameters $\theta_\text{ref}$; initial state distribution $\mathcal{D}_0$; Group size $G$; learning rate $\eta$; Hyperparameters $\epsilon$, $\beta$; Sampled group size $G'$' Sampling temperature $T_\text{ep}$
\STATE Initialize policy parameters $\theta \leftarrow \theta_\text{ref}$
\FOR{training iteration $= 1, \ldots, M$}
    \STATE Set $\theta_\text{old}\leftarrow\theta$
    \STATE Sample initial state $s_0 \sim \mathcal{D}_0$
    \STATE Generate initial observation $o_0$ from $s_0$
    \FOR{episode $j = 1$ to $G$}
        \STATE Set $o_{j,0}\leftarrow o_0$
        \FOR{episode step $t = 0$ until termination}
            \STATE $q_{j,t}=\mathcal{T}(o_{j,t})$ \COMMENT{Construct prompt}
            \STATE $y_{j,t}\sim p_{\theta_\text{old}}(\cdot|q_{j,t})$ \COMMENT{Generate completion}
            \STATE $a_{j,t}=\psi(y_{j,t})$ \COMMENT{Parse action}
            \STATE Take action $a_{j,t}$, observe $o_{j,t+1}$ and $r_{j,t+1}$
            \IF{terminal state} \STATE break inner loop
            \ENDIF
        \ENDFOR
        \STATE Compute reward $C_j = \sum_{t}(r_{j,t+1} + \Phi_j,t)$
    \ENDFOR
    \STATE Compute advantages $\{A_i\}_{j=1}^G$ as normalized rewards
    \IF{$T_\text{ep}>0.0$}
        \STATE Sample $G'$ episodes using AAW Sampling
        \STATE Recompute $\{A_i\}_{j=1}^{G'}$ using only sampled episodes
    \ENDIF
    \STATE Update policy parameters using gradient ascent: $\theta \rightarrow \theta + \eta \nabla_\theta \mathcal{J}_{\text{MS-GRPO}}(\theta)$
\ENDFOR
\RETURN $\pi_\theta$
\end{algorithmic}
\end{algorithm}

\subsection{Absolute-Advantage-Weighted Episode Sampling}
\label{subsec:episode_sample}
To improve training efficiency, we propose Absolute-Advantage-Weighted (AAW) episode sampling. This strategy prioritizes episodes with high-magnitude advantages, inspired by Prioritized Experience Replay \citep{schaul_prioritized_2016}. The intuition is that these episodes, representing the most significant success or failures, are the most informative for learning.

We calculate the group relative advantage (Equation \ref{eqn:advantage}) across all $G$ generated episodes, then sample $G'< G$ episodes without replacement. The probability of selecting episode $j$ is given by the Softmax over the scaled absolute advantages:
\begin{equation}
p_j = \frac{\exp(|A_j|/T_\text{ep})}{\sum_{i=1}^{G} \exp(|A_i|/T_\text{ep})}
\end{equation}
where the temperature $T_\text{ep}\in(0,\infty)$ controls the strength of the weighting. Smaller $T_\text{ep}$ concentrates sampling on extreme advantage episodes whereas larger $T_\text{ep}$ approaches a uniform distribution.




\section{Experimental Setup}
We evaluate our proposed methodology through a series of experiments. Our experiments aim to determine if MS-GRPO can improve the decision-making capabilities of a small LLM and to assess whether those improvements generalize to unseen environments or variants of the training environment. To achieve this, we use two 2D grid-world environments, Snake and Frozen Lake, and evaluate agent performance using the total cumulative reward per episode.

\subsection{Environments}
\label{subsec:environments}
We choose Snake and Frozen Lake because their dynamics are simple yet challenging for small language models, making them ideal for assessing the learning algorithm's effectiveness. Their simple structure allows for creating variants to test generalization.

Both environments have identical action spaces, $\mathcal{A}=\{Up, Down, Left, Right\}$, and similar objectives, each requiring the agent to navigate a 2D grid towards a goal while avoiding dangers. For a LAP agent, solving these tasks requires identifying goals and dangers from a text observation, planning a strategy, and faithfully translating that plan into actions. The recovery strategy for an invalid action is to take no action.

\paragraph{Snake} The agent controls a snake that grows longer by consuming fruit. A non-LAP agent controls a second snake, which takes random valid actions (avoiding walls and its own tail). Episodes terminate on collision with the snake's own tail, another snake, or the grid boundaries. Fruit is replaced in a random empty board tile when consumed. This environment is adapted from \citet{greg_kamradt_snake_2025}.

\paragraph{Frozen Lake}
The agent navigates a grid of ice tiles to reach a goal. Some tiles contain holes, and moving on to one terminates the episode. A safe path to the goal is guaranteed to exist and moving into a wall has no effect. The environment is from the Gymnasium library \citep{towers_gymnasium_2024}.

\paragraph{Environment Variants}
We create variants of each environment to test different aspects of generalization:

\begin{itemize}
    \item \textbf{Snake-Standard} (training/evaluation): A 10x10 grid with one other snake and 5 apples giving $+1$ reward. A collision results in $-3$ reward and terminates the episode.
    \item \textbf{Snake-Poison} (evaluation): Like Snake-Standard, but apples provide $-1$ reward, testing the agents ability to override its training objective.
    \item \textbf{FrozenLake-NotSlippery} (training/evaluation): A 4x4 grid where each tile has $0.2$ probability of being a hole. Reaching the goal gives +1 reward.
    \item \textbf{FrozenLake-Slippery} (evaluation): like FrozenLake-NotSlippery, but movement is stochastic. The agent moves in the chosen direction with $1/3$ probability and a perpendicular direction with 1/3 probability each. This variant tests planning under uncertainty.
\end{itemize}

\subsection{Agent-Environment Interface}
Each environment's state is converted to text by the observation function $O$ of the TMSG. We provide the observation in two ways concurrently: (1) as a list of entity coordinates and (2) as a 2D character grid. These are supplemented with static text describing their meaning. A static description of the environment's rules and goals is prepended to the dynamic state representation to form the observation, $o$. The static and dynamic text for each environment variant is presented in the technical appendix. The complete observation is inserted into the LAP agent's template $\mathcal{T}$, which provides environment-agnostic instructions on reasoning structure and output formatting.

\subsection{Reward Design}
The reward signal guides the agent towards two objectives: maximizing its environment reward and generating well-formatted text. While the environment reward alone may implicitly encourage good formatting, we add an explicit format penalty, $\Phi$, an approach followed by \citet{deepseek-ai_deepseek-r1_2025}. The agent's total reward is a composite of two components:
\begin{itemize}
    \item \textbf{Environment Reward ($R$):} The native reward from the environment, plus a $-0.5$ penalty per invalid action.
    \item \textbf{Format Penalty ($\Phi$):} A set of penalties for undesirable text patterns:
    \begin{itemize}
        \item \textit{Length Penalty}, a linear penalty for excessive text generation, scaling from $0$ to $-0.5$ for responses between $180$ and $200$ tokens
        \item \textit{Structure Penalty}, a $-0.5$ penalty for each missing, unnecessary, or incorrectly nested XML tag
        \item \textit{Extra Text Penalty}, a $-0.5$ penalty if any text is generated after the final $</\text{action}>$ tag
\end{itemize}
\end{itemize}

\subsection{Experimental Protocol and Models}

We use the \textit{Qwen2.5-3B-Instruct} \citep{qwen_qwen25_2025} model for post-training as its size offers a balance between capability and computational footprint. We compare its post-trained performance against two larger models, \textit{Qwen2.5-32B-Instruct} and \textit{Qwen2.5-72B-Instruct}. We also train a Deep Q-Network (DQN) \citep{mnih_playing_2013} on \textit{Snake-Standard} to provide a non-LLM baseline. Agents are trained separately on the \textit{Snake-Standard} and \textit{FrozenLake-NotSlippery} environments and evaluated on all four variants. We conduct an ablation study  comparing the effectiveness and time efficiency of MS-GRPO with and without AAW sampling. Full training parameters, evaluation details, and LAP definitions are provided in the technical appendix.

\section{Results}

Our experiments demonstrate that MS-GRPO can successfully improve the sequential decision-making capabilities of LLMs. This section highlights several key findings: post-training improves performance on both training environments, but with high variance on Snake; our post-trained 3B model outperforms a 72B baseline on Frozen Lake; and our AAW strategy shows signs of improving performance without sacrificing time efficiency. In addition we find that a DQN trained baseline vastly outperforms our agents, and see mixed evidence of generalization.


\subsection{Post-training with MS-GRPO Improves Sequential Decision-Making}

\begin{figure}[!htb]
    \centering
    \includegraphics[width=1.0\linewidth]{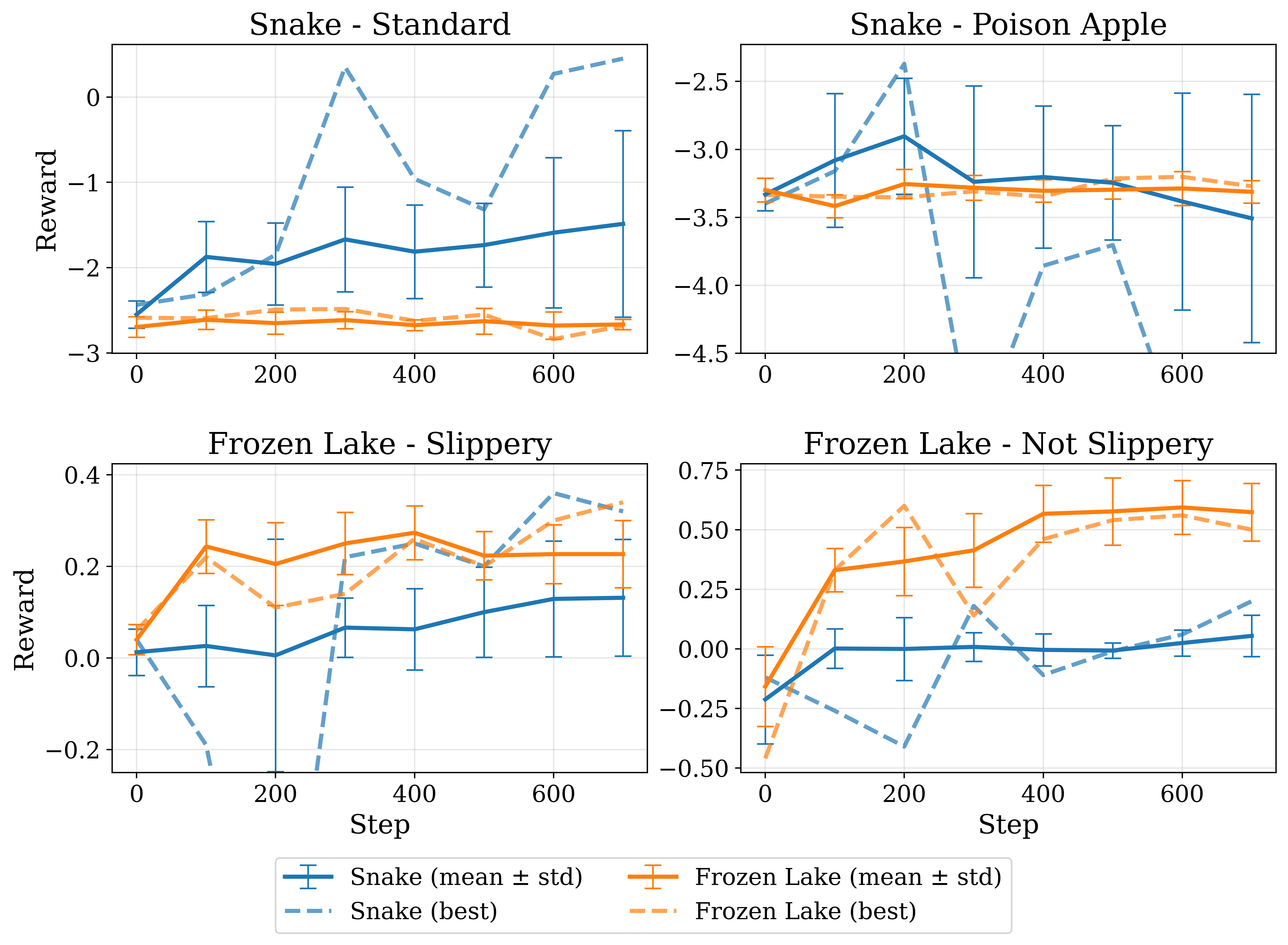}
    \caption{Training progression over 700 steps across four evaluation scenarios. The solid line in each panel shows the mean reward across 8 independent training runs, with the error bars representing the standard deviation. The dashed line shows the performance of the single best-performing run. At each step, performance is the mean reward over the same 50 evaluation episodes.}
    \label{fig:training_curve}
\end{figure}

Post-training with MS-GRPO leads to significant performance improvements on the agents' respective training environments, demonstrated by the upward trend of the learning curves in Figure \ref{fig:training_curve}. As shown in Table \ref{tab:learning_curves} both agents improve on their native task, but there is high variability in the final performance of the Snake-trained agents.

For example, the best run for a Snake-trained agent achieves a reward of $0.45$ on the \textit{Snake-Standard} evaluation, considerably greater than the mean of $-1.49$ and demonstrating the high degree of variance in the training outcomes. This variance indicates that the training process is sensitive to initial conditions or early exploration, with some agents converging on effective strategies while others stagnate.

\begin{table*}\centering
\caption{Performance comparison across environments at initial and final training steps, with per-run difference statistics. Values shown as mean (std) of the evaluation environment reward.}
\label{tab:learning_curves}
\begin{tabular}{llrrr}
\toprule
Training & Evaluation & Initial - 0 & Final - 700 & $\Delta$ \\
\midrule
\multirow[t]{6}{*}{\textbf{Snake}} & \textbf{Snake - Standard} & -2.607 (0.162) & -1.487 (1.093) & +1.120 (1.001) \\
\textbf{} & \textbf{Snake - Poison Apple} & -3.298 (0.126) & -3.508 (0.913) & -0.210 (0.827) \\
\textbf{} & \textbf{Frozen Lake - Slippery} & 0.020 (0.035) & 0.131 (0.127) & +0.111 (0.114) \\
\textbf{} & \textbf{Frozen Lake - Not Slippery} & -0.207 (0.213) & 0.054 (0.087) & +0.261 (0.203) \\
\cline{1-5}
\multirow[t]{6}{*}{\textbf{Frozen Lake}} & \textbf{Snake - Standard} & -2.696 (0.120) & -2.665 (0.061) & +0.030 (0.143) \\
\textbf{} & \textbf{Snake - Poison Apple} & -3.299 (0.086) & -3.312 (0.083) & -0.013 (0.049) \\
\textbf{} & \textbf{Frozen Lake - Slippery} & 0.040 (0.033) & 0.227 (0.073) & +0.187 (0.059) \\
\textbf{} & \textbf{Frozen Lake - Not Slippery} & -0.158 (0.167) & 0.573 (0.121) & +0.732 (0.201) \\
\cline{1-5}
\bottomrule
\end{tabular}
\end{table*}

\subsection{Post-trained Model Outperforms Larger Baselines}
\begin{figure}[!htb]
    \centering
    \includegraphics[width=1.0\linewidth]{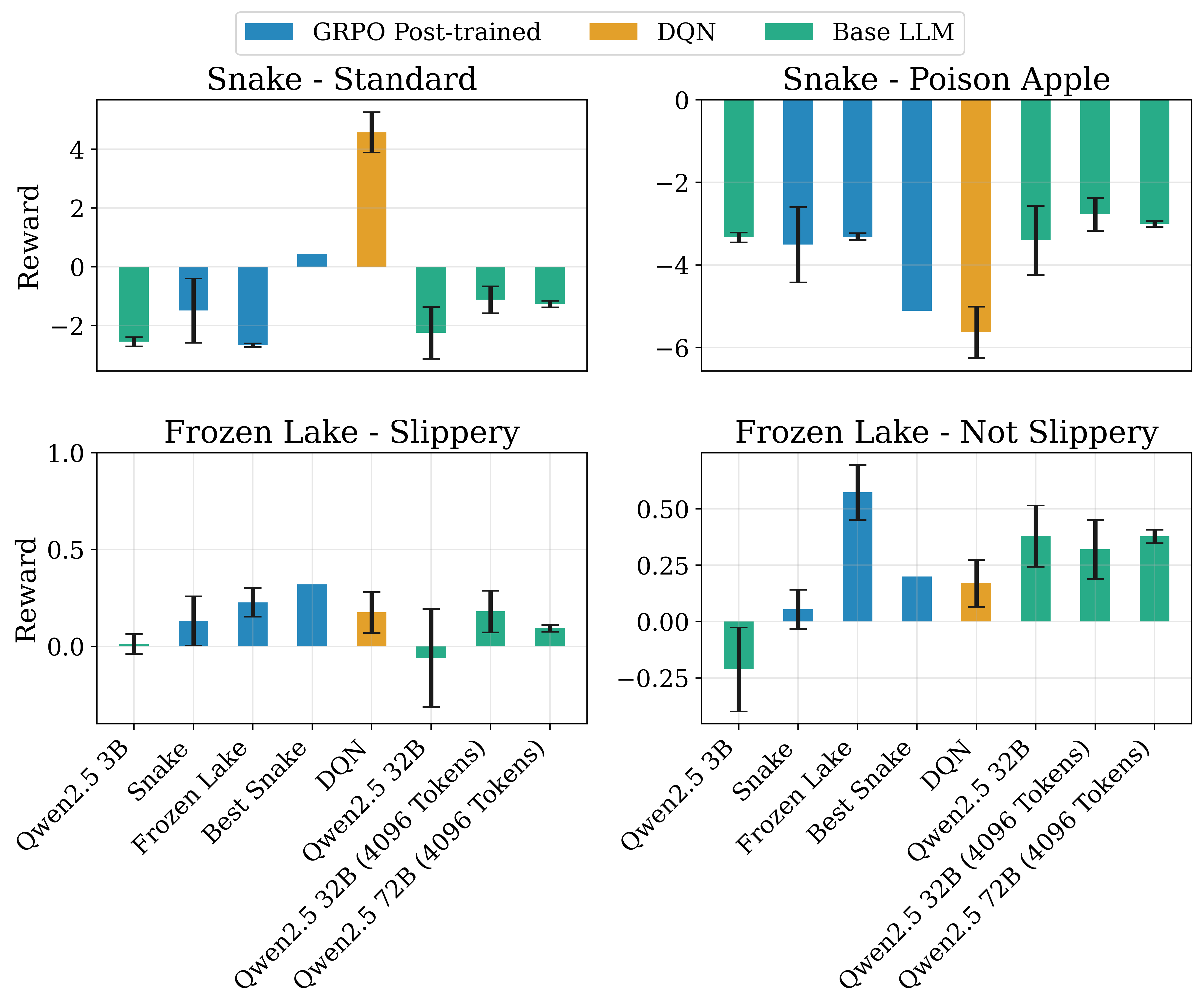}
    \caption{Mean reward per episode for MS-GRPO post-trained agents and baselines across four evaluation environments. MS-GRPO results are averaged over 8 training runs, each evaluated on $50$ episodes. Baselines are evaluated on $1,000$ episodes. Error bars show standard deviation across runs for MS-GRPO agents and $95\%$ confidence intervals for baselines.}
    \label{fig:baseline_comparisons}
\end{figure}

A direct comparison shows that our post-training method enables the 3B parameter model to outperform its much larger counterparts.
As illustrated in Figure \ref{fig:baseline_comparisons}, our Frozen Lake post-trained agent achieves $0.57\pm 0.12$ on its training environment, \textit{FrozenLake-NotSlippery}, surpassing the $0.38\pm0.48$ achieved by the 72B parameter model, despite operating with a 200-token limit compared to the baseline's 4096-token limit.

On the other hand, the mean reward of our post-trained Snake agents showed no clear improvement over the larger LLMs. However, the single best performing \textit{Snake-Standard} agent achieved a final reward of $0.45$ on \textit{Snake-Standard} compared to $-1.26\pm 1.80$ for the 72B model, and $0.32$ on \textit{FrozenLake-Slippery} compared to $0.094\pm0.29$.

These findings demonstrate that task-specific post-training can be more practical and efficient than scaling model size for sequential decision-making tasks.

\subsection{DQN Outperforms MS-GRPO on In-Domain Task}
The DQN agent significantly outperforms even the best single MS-GRPO on \textit{Snake-Standard} as shown in Figure \ref{fig:baseline_comparisons}. It achieves $4.58\pm 2.47$ compared to $0.45$ by the best MS-GRPO snake agent and $-1.49\pm 1.09$ mean across all snake agents. 
This performance gap highlights the challenges in using general-purpose language models for specific tasks that are poorly represented in their pre-training data. To maximize performance on a specific task, a specialized model is superior.

\subsection{MS-GRPO Post-trained Snake Agent Generalizes to an Unseen Frozen Lake Environment}

The best performing Snake-trained agent shows promising zero-shot generalization to an unseen task, achieving a higher mean reward ($0.32$) on the \textit{FrozenLake-Slippery} task than the DQN agent ($0.17\pm0.38$). This suggests the MS-GRPO agent, despite its substantially worse performance on the Snake environment, is at adaptin to novel dynamics. 

However, the same Snake agent's generalization performance on the \textit{Snake-PoisonApple} task degraded after post-training, becoming considerably worse than the base model it originated from (Figure \ref{fig:baseline_comparisons}). This suggests that the agent's learned behavior for seeking apples cannot be offset by instructions in the prompt stating that they are poisoned. 


\subsection{AAW Sampling Improves Performance Without Impacting Training Time}

\begin{figure}[!htb]
    \centering
    \includegraphics[width=1.0\linewidth]{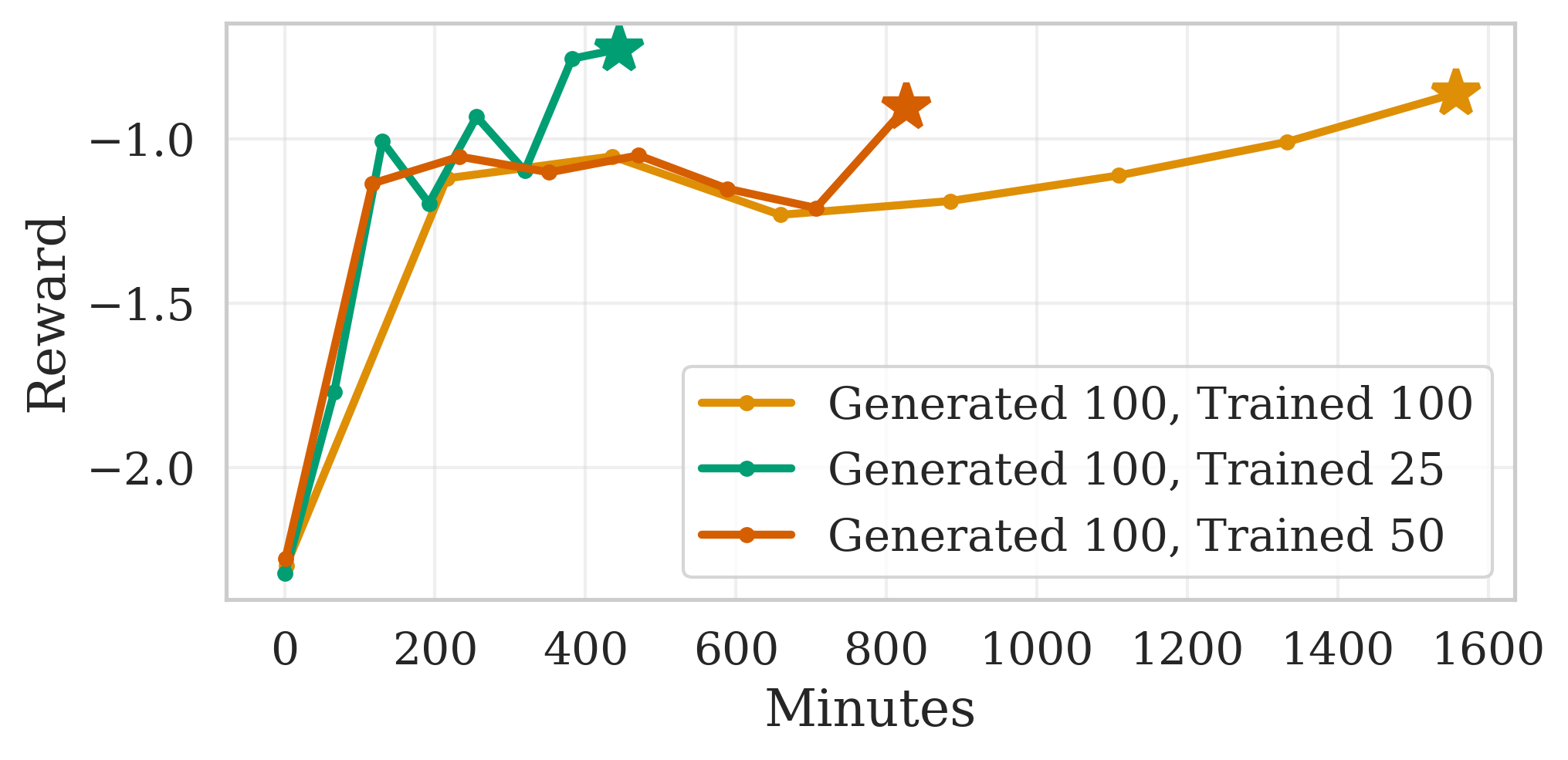}
    
    \vspace{0.1cm}

    \includegraphics[width=1.0\linewidth]{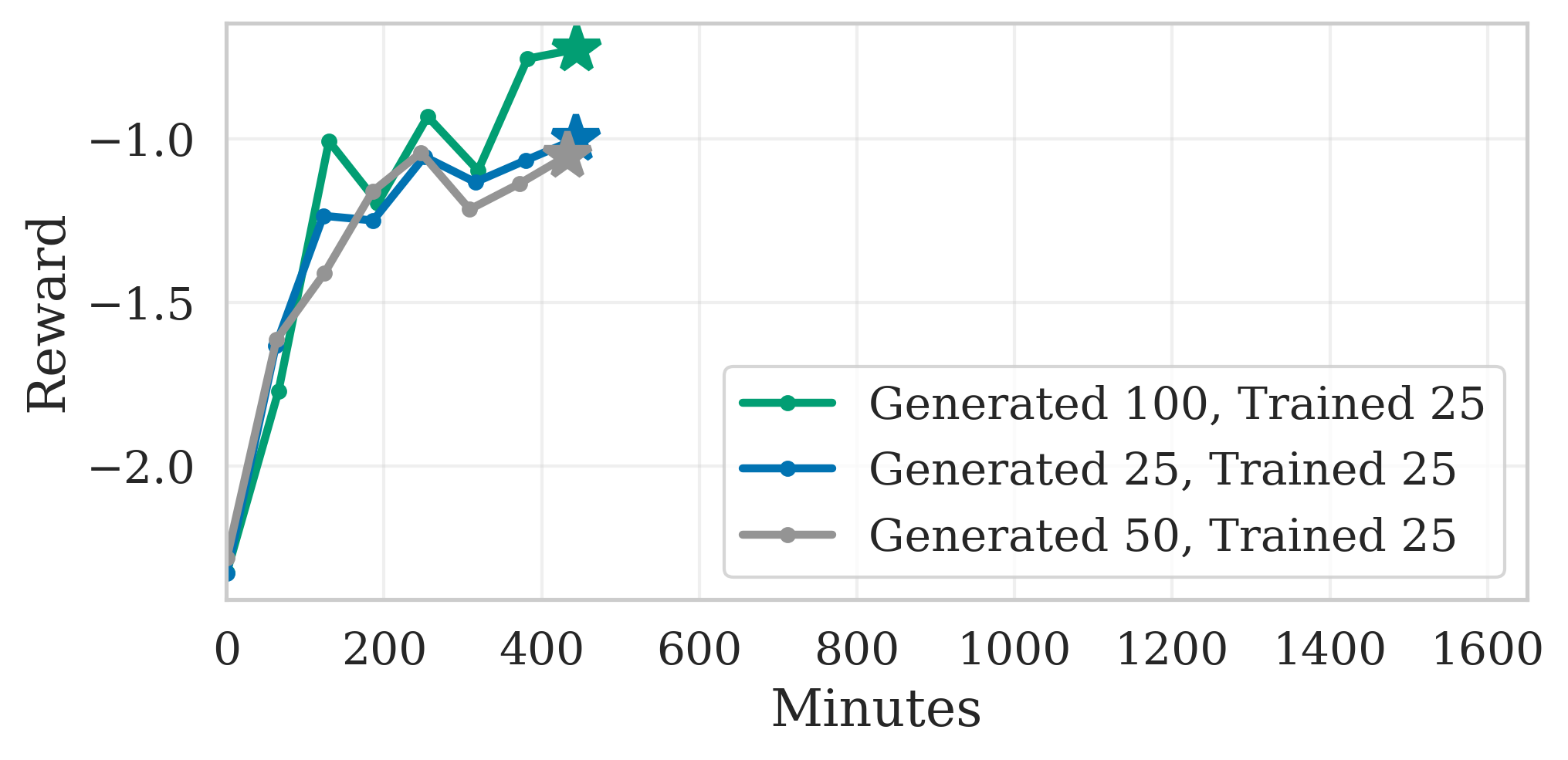}
    
    \caption{Ablation study on convergence and training efficiency with various degrees of AAW sampling, showing mean reward versus wall time for different sampling configurations on the \textit{Snake-Standard} evaluation task.
    \textbf{(Top)} Varying the number of sampled episodes $G'$ for a fixed number of generated episodes $G=100$.
    \textbf{(Bottom)} Varying the number of generated episodes for a fixed number of sampled episodes $G'=25$}.
    \label{fig:sampled_curves}
\end{figure}

We find that our AAW sampling strategy reduces training time while maintaining or improving performance. As shown in Figure \ref{fig:sampled_curves}, training with $G=100$ and $G'=25$ over 700 steps provides 3.5x time savings compared to the unsampled baseline ($G=100$, $G'=100$), while achieving comparable final rewards ($-0.72$ vs. $-0.86$). At the same time, by generating additional episodes for a fixed number of training episodes, we see a higher reward: $-0.72$ with $G=100$ and $G'=25$, compared to $-1.00$ with $G=25$ and $G'=25$. This suggests that, when training on the Snake environment, our sampling strategy successfully selects higher-quality episodes without substantially increasing computational load.

These results indicate that for our particular environment and model, using AAW can improve both efficiency and performance. 


\section{Discussion}

Our experimental results demonstrate that MS-GRPO can successfully post-train language models on sequential decision-making tasks. However, the results also reveal considerable limitations in training consistency and a significant performance gap relative to a bespoke DQN agent. The large performance difference between the best Snake agent and the mean highlights both the potential of the training method and the need to improve its consistency.

We hypothesize that this inconsistency stems from insufficient exploration during training, where finding an effective policy is left too much down to initial conditions and chance. Unlike traditional RL agents that directly explore the state-action space, helped by methods such as $\epsilon$-greedy sampling, exploration for LAP agents is the indirect result of exploring the token space. Our LAP framework makes this problem explicit by defining the distinct components that can be used to control agent behavior. Additionally to the LLM parameters, the agent's generation configuration, $\mathcal{G}_i$, and prompt template, $\mathcal{T}_i$, provide means for influencing how the LAP takes actions. For example, dynamically adapting the text sampling temperature in $\mathcal{G}_i$ to increase when responses or rewards stagnate could enable the learning algorithm to adapt, so that text generation never becomes too consistent during training, a prerequisite for environment exploration. Alternatively, training the agent with a varied set of prompt templates ($\mathcal{T}_i$) to elicit a variety of behaviors may also promote more thorough exploration of the environment. 

Another possible cause of inconsistent training is the use of an imprecise Monte Carlo credit assignment scheme, which may dilute the learning signal from truly effective behaviors that the agent explores.

Furthermore, the failure of the best performing Snake agent on the \textit{Snake-PoisonApple} task highlights a risk of the post-training process: reinforcing a specific skill may prevent the model from addressing critical semantic details relevant to that skill. While the training successfully enhanced the agents ability to seek apples, it was not able to correctly adapt its behavior to the scenario in which apples are described as poisonous, with performance degrading relative to the base model as a result.


Our AAW sampling approach shows promising performance gains (all three experiments using $G=100$ outperformed those with smaller $G$) without sacrificing time efficiency. However, the improvements are modest and would benefit from further validation across more environments. 

Finally, the improved performance over larger LLMs demonstrates the value of task-specific post-training of smaller LLMs. Using smaller models with fewer generated tokens reduces computational requirements and improves response times, making the model more practical for real-world applications. However, the performance gap relative to a specialized DQN agent highlights a fundamental limitation of this approach. Even with clear improvement over the base models and signs of generalization, the agent's absolute performance on a narrow, well-defined task falls short of what simpler, specialized alternatives can achieve. This suggests that the value of LLM-based agents may not be their ability to outperform specialized agents, but rather their flexibility to tackle the wide range of scenarios that an agent might encounter in the real-world scenarios.





\section{Conclusions}

In this work, we investigated whether the decision-making capabilities of small LLMs could be improved for sequential decision-making tasks without relying on extensive reasoning chains. To this end, we introduced the Multi-Step Group-Relative Policy Optimization (MS-GRPO) post-training algorithm. Our experiments serve as a demonstration that this approach is effective: a post-trained 3B parameter model outperformed a 72B parameter baseline, showing that targeted training can be a more effective route to improving capability than scaling model size. Additionally, we tested a selective episode sampling strategy and found indications that it improves task performance without impacting training time efficiency. This work establishes a methodology for creating more efficient and practical LLM-based decision-making agents. 


Our findings point to two key directions for future work. First, our use of a simple Monte Carlo credit assignment mechanism likely contributes to the observed training inconsistency. Exploring more nuanced approaches could help provide a more precise learning signal to improve performance. Second, while our agent demonstrated promising zero-shot generalization to a novel environment's dynamics, the simultaneously failure on the semantically simple poisoned apple scenario highlights a critical challenge: ensuring that post-training does not override the model's core semantic reasoning capabilities. Addressing these challenges will be crucial for enabling practical LLM-agents that are not only efficient, but also robust and adaptable.

\bibliography{references}


\section*{Supplementary material}

# Technical Appendix

## Agents

**LAP Agents**

We used *Qwen2.5-3B-Instruct* as the base model for post-training in all experiments, using Low-Rank Adaptation (LoRA) targeting parameter updates on all linear layers (Hu et al., 2021). LAP definitions for this and the comparative base LLMs are provided in Table 1. All agents have the same template $\mathcal{T}$ and action parser $\psi$.

Table 1: LAP definitions for the agents used in the experiments. The prompt template $\mathcal{T}$ and action parser $\psi$ are the same for all agents.

| Agent | LLM ($\mathcal{L}_\theta$) | Generation Configuration ($\mathcal{G}$) |
|---|---|---|
| $\Pi_{\text{MS-GRPO}}$ | *Qwen2.5-3B-Instruct* + MS-GRPO LoRA updates | **Training:** Temp=1.5, top-k=3, tokens=200 <br> **Evaluation:** Greedy (Temp=0), tokens=200 |
| $\Pi_{\text{3B}}$ | *Qwen2.5-3B-Instruct* | Greedy (Temp=0), tokens=200 |
| $\Pi_{\text{32B}}$ | *Qwen2.5-32B-Instruct* | Greedy (Temp=0), tokens=200 |
| $\Pi_{\text{72B-200}}$ | *Qwen2.5-72B-Instruct* | Greedy (Temp=0), tokens=200 |
| $\Pi_{\text{72B-4096}}$ | *Qwen2.5-72B-Instruct* | Greedy (Temp=0), tokens=4096 |

**Action Parser $\psi$**    An action string is extracted from the LLM's response at each timestep. The action parsing function extracts the text inside the first set of < action > $\cdots$ < /action > tags. Each environment implementation must define a mapping from text string to action index. The action index corresponding to the extracted string is used as the agent's action in the next timestep. If there is not a valid pair of action tags, or the extracted string is not in the mapping, no action is taken.

**Template $\mathcal{T}$**    The LAP agent template, presented below, is shared for all LAP agents used in this study. There are two slots into which the observation is inserted. The static part of the observation, containing environment specific rules, is inserted into `{environment_prompt}`. The dynamic part of the observation, containing a combination of dynamic text describing the environment state and static template explaining the representation of the state, is inserted into `{game_state}`.

## DQN Agent

We trained a Deep Q-Network (Mnih et al., 2013) to act as a comparison in our experiments. We used a convolutional neural network with architecture detailed in Table 2, using a Rectified Linear Unit (ReLU) activation function after each convolutional and hidden fully-connected layer. We also performed a parameter sweep to determine the training hyperparameters. The search space is as follows:

**Learning Rate** ($\alpha$) over $\{10^{-5}, 10^{-4}, 10^{-3}\}$

**Discount Factor** ($\gamma$) over $\{0.9, 0.95, 0.99\}$

**$\epsilon$-greedy Decay Steps** over $\{5 \times 10^3, 2 \times 10^4, 10^5, 5 \times 10^5, 10^6\}$

The model presented for comparison in the Results, as determined by the greatest evaluation environment reward, used $\alpha = 10^{-5}$, $\gamma = 0.9$, $\epsilon_{\text{decay}} = 10^5$. The following hyperparameters were used for all DQN training experiments: initial $\epsilon = 1.0$, final $\epsilon = 0.1$, replay buffer size of $10^4$, batch size of 128, and a target network update frequency of 1,000 steps. All experiments used 6 million training episodes.

# Experiment Configuration

## Training Protocol

We trained two types of agent: one exclusively on the *Snake-Standard* environment and another on *FrozenLake-Standard*. When training, we limited the number of episode steps to 10. We repeated training 8 times with different random seeds. All agents were then evaluated on all four environment variants.

Training was conducted for 700 steps of Algorithm 1 from the Methodology. We generated a group of $G = 100$ episodes and sampled $G' = 25$ from those, with an episode sampling temperature of

Table 2: Overview of the DQN architecture structure. Input shape is $(B, C, H, W)$, where $B$ is the batch size, $C$ is the number of input channels, $H$ is the height of the environment grid and $W$ is its width. $A$ is the number of discrete actions.

| Layer Block | Layer Type | Parameters / Details | Output Shape |
|---|---|---|---|
| **Input** | - | - | $(B, C, H, W)$ |
| Conv 1 | Conv2d | 32 filters, kernel 3x3, stride 1, pad 1 | $(B, 32, H, W)$ |
| | ReLU | - | $(B, 32, H, W)$ |
| Conv 2 | Conv2d | 64 filters, kernel 3x3, stride 1, pad 1 | $(B, 64, H, W)$ |
| | ReLU | - | $(B, 64, H, W)$ |
| Conv 3 | Conv2d | 64 filters, kernel 3x3, stride 1, pad 1 | $(B, 64, H, W)$ |
| | ReLU | - | $(B, 64, H, W)$ |
| **Flatten** | - | - | $(B, 64 \times H \times W)$ |
| FC 1 | Linear | 512 output units | $(B, 512)$ |
| | ReLU | - | $(B, 512)$ |
| **FC 2 (Output)** | Linear | $A$ output units | $(B, A)$ |

$T_{\text{ep}} = 0.1$. In each episode, a maximum of 5 LAP actions (each of which can specify multiple sequential environment actions) and 10 environment steps was allowed. The LLM sampling temperature was 1.5 and used top-k sampling with $k = 3$. Other hyperparameters for MS-GRPO were a learning rate of $1 \times 10^{-4}$, clipping values of $\epsilon_{\text{low}} = \epsilon_{\text{up}} = 0.1$ and a KL-penalty weight of $\beta = 0.1$.

---

*FrozenLake-NotSlippery* | **Static** Observation Text | {environment_prompt}

```
You are navigating the surface of a frozen lake. You must reach the goal.
Rules:
If you step on a hole, you will fall through and die.
Your available actions are: Up, Down, Left, Right. You can make between 1 and 3
    ↪ actions, separated by the action separator " || "
```

---

*FrozenLake-NotSlippery* | **Dynamic** Observation Text Example | {game_state}

```
The board size is 4x4. Normal (X, Y) coordinates are used ranging from.
LEFT decreases X, RIGHT increases X, UP increases Y, and DOWN decreases Y.
Coordinates range from (0, 0) at bottom left to (3, 3) at top right.
Player position: (0, 3)
Holes: (1, 3), (2, 3), (3, 3), (3, 2)
Goal: (3, 0)
The meaning of each symbol in the state is:
- P: Player
- O: Hole
- G: Goal
- _: Empty space
State:
P O O O
_ _ _ O
_ _ _ _
_ _ _ G
```

## Evaluation Protocol

To provide a comparison for the performance of the post-trained models, we used two larger models without MS-GRPO post-training, *Qwen2.5-32B-Instruct* and *Qwen2.5-72B-Instruct*. For the 72B parameter model, we evaluated once with a maximum number of generated tokens equal to that of the post-trained models (200) and once with a much greater limit (4096), providing both a like-for-like comparison as well as a measure of the model's full capability. Environment variant specific settings are detailed in the Experimental Setup section of the paper. The dynamic observation texts are identical to those in the Training Protocol above. The static observation texts for *Snake-PoisonApple* and *FrozenLake-Slippery* are detailed below. During evaluation, a longer episode of 20 LAP actions and environment steps was allowed to better assess long term performance. We used greedy decoding (text generation sampling temperature= 0). For consistency, the evaluation configuration file seed is set to 0 for all experiments, ensuring that we always use the same set of randomly generated initial conditions for evaluation.

```
Snake-PoisonApple | Static Observation Text | {environment_prompt}


You are controlling a snake in a multi-player Snake game
Rules:
- You can move your head one space up, down, left, or right
- If you move onto an apple, you *lose* 1 point. You must avoid the apples for
  ↪ as long as possible.
- You die if you move into a wall, another snake, or yourself
Your available actions are: Up, Down, Left, Right. You can make between 1 and 3
  ↪ actions, separated by the action separator " || "
```

```
FrozenLake-Slippery | Static Observation Text | {environment_prompt}


You are navigating the surface of a frozen lake. You must reach the goal. If you
  ↪ step on a hole, you will fall through and die. You may move in an
  ↪ unintended direction due to the slippery ice, including into a hole.
Your available actions are: Up, Down, Left, Right. You can make between 1 and 3
  ↪ actions, separated by the action separator " || "
```

**Absolute-Advantage-Weighted (AAW) Episode Sampling Ablation Study**

To evaluate the effectiveness of our AAW sampling strategy, we conducted two sets of experiments. First, we generated $G = 100$ episodes per training step and compared training on different subset sizes ($G' = 25$, 50 or 100 episodes). Additionally, we varied the total number of generated episodes ($G = 25$, 50, 100) and kept the training subset fixed at $G' = 25$ episodes. All subsets were sampled using $T_{ep} = 0.1$. All experiments used identical hardware, detailed in the Hardware section below. Other training settings were identical to those used in Training Protocol.

# Hyperparameter Selection

When analyzing the MS-GRPO algorithm, we performed parameter sweeps for LLM generation temperature in the range 0.2 to 2.0 and top-k in the range $k = 1$ to $k = 10$ and without a limit. We determined the selected values for our experiments based on the evaluation reward on *Snake-Standard* for agents trained on Snake. We found the combination Temperature= 1.5 and $k = 3$ to give the best mean evaluation reward over 3 runs. Similarly, we used the results from the episode sampling study to determine which values of $G$ and $G'$ to use, finding that $G = 100$ with $G' = 25$ gave the best combination of training time efficiency and evaluation reward on *Snake-Standard*.

# Hardware

All training and evaluation was performed on one of two types of hardware:

- *NVIDIA A100-SXM4-80GB* graphics card with AMD EPYC 7413 24-Core rocessor

- *NVIDIA H100 80GB HBM3* graphics card with Intel Xeon Platinum 8468 48 Core processor

Both setups use *Red Hat Enterprise Linux 8.9 (Ootpa)*. MS-GRPO training used a single GPU of either configuration. Evaluation of $\Pi_{32B}$ used 2 H100 GPUs, linked by NVLink, and $\Pi_{72B}$ and $\Pi_{72B-4096}$ used 4 H100 GPUs, linked by 2x NVLink and combined with 1x NVSwitch.

All training and evaluation for the AAW sampling experiments used the H100 configuration.

 
\end{document}